\documentclass[11pt]{article}
\usepackage[margin=1in]{geometry}
\usepackage[T1]{fontenc}
\usepackage{booktabs}
\usepackage{amsmath}
\usepackage{array}
\usepackage{graphicx}
\usepackage{makecell}
\usepackage{multirow}
\usepackage{algorithm}
\usepackage{algpseudocode}
\usepackage{url}
\usepackage[colorlinks, urlcolor=blue]{hyperref}
\usepackage{xcolor}

\begin{document}

\title{Lightweight Polyp Segmentation via a Gain-Aware Prediction-Space Recursive Controller}

\author{
Jiachi Zhang$^{1,3}$, Zhuoyu Wu$^{2}$, Quanjun Wang$^{3}$, Wenhui Ou$^{4}$, and Wenqi Fang$^{1}$\\
\small $^1$Shenzhen Institutes of Advanced Technology, Chinese Academy of Sciences, Shenzhen, China\\
\small $^2$CyPhi($\Psi\Phi$) AI Research Lab, School of IT, Monash University Malaysia\\
\small $^3$South China University of Technology, Guangzhou, China\\
\small $^4$Department of Electronic \& Computer Engineering, The Hong Kong University of Science and Technology\\
\small Corresponding author: Wenqi Fang\\
\small \texttt{wq.fang@siat.ac.cn}
}
\date{}

\maketitle

\begin{abstract}
While lightweight polyp segmentation is highly desirable for low-cost deployment, reported performance gains often stem from upgraded backbone encoders, complex decoders, or heavy refinement branches. Consequently, it remains difficult to isolate whether a lightweight correction mechanism is inherently effective on its own. We address this limitation by formulating refinement as a prediction-space recursive correction task, introducing a recursive controller that operates directly on backbone logits. Under a fixed recursion budget, this controller aggregates discrepancy and uncertainty evidence, updates a compact state tracking recent correction utility, and applies additive residual logit corrections. By design, this correction path remains small, host-portable, and deployment-explicit. Utilizing a unified Kvasir-trained protocol, we evaluate our approach across seven lightweight backbones on Kvasir-SEG and three transfer datasets, measuring segmentation accuracy (Dice/IoU) alongside deployment efficiency (parameters, GMACs, and peak memory). The controller yields consistent improvements in the source domain, achieves competitive performance against both training-side baselines and heavier structural refiners on representative hosts, and delivers selective transfer gains with minimal static overhead.
Code is available at \url{https://github.com/tyui99/Gain-Aware-Prediction-Space-Recursive-Controller}.


\end{abstract}

\section{Introduction}
Colonoscopy is the gold standard for colorectal cancer screening, demanding accurate, real-time polyp segmentation for the precise detection and resection of precancerous polyps~\cite{tesema2026lgps}. To achieve an optimal trade-off between efficiency and performance, lightweight segmentation models offer a highly practical solution for real-world clinical deployment~\cite{zami2026ultralightcps}. However, these models still degrade significantly when confronting ambiguous boundaries, small lesions, and noisy regions~\cite{yue2024boundary}. To mitigate these limitations, recent approaches predominantly focus on reinforcing the representation pathway by integrating heavier encoders~\cite{ramos2026multi}, intricate decoder interactions~\cite{zhao2026dgfi}, or auxiliary refinement branches~\cite{wu2024harmonizing}. While effective, this strategy fundamentally alters the base network itself, preventing performance gains from being cleanly attributed to an independent, lightweight correction mechanism.

This attribution problem carries both scientific and practical implications; once performance gains become entangled with an augmented feature pathway, the correction mechanism becomes difficult to isolate, less transferable across different host architectures, and challenging to evaluate under strict deployment constraints. Consequently, we investigate the feasibility of designing a compact, plug-and-play module that enhances lightweight segmentation by directly refining predictions, thereby avoiding the need to re-engage a heavy feature hierarchy.

In summary, our key contributions are two-fold:
\begin{itemize}
\item We propose a lightweight, model-agnostic segmentation framework that pairs a flexible single-pass backbone with a compact, three-stage recursive controller. This design shifts the refinement process entirely into the prediction space, utilizing discrepancy and uncertainty info to iteratively correct anchor logits over $T$ iterations without re-engaging heavy feature hierarchies.

\item We conduct extensive benchmarking against various standard backbones and competitive logit-refinement mechanisms, demonstrating the clear performance and architectural advantages of our approach and validating its efficacy in enhancing lightweight segmentation models.

\end{itemize}

\section{Related Work}
Lightweight polyp segmentation traditionally improves performance by optimizing feature pathways, evolving from the classical U-Net~\cite{ronneberger2015u} to hybrid encoder-decoder variants like TransUNet~\cite{chen2021transunet}. Representative lightweight hosts—such as ULite~\cite{dinh20231m}, Mobile-PolypNet~\cite{karmakar2022mobile}, CGNet~\cite{wu2020cgnet}, EGE-UNet~\cite{ruan2023ege}, CMUNeXt-S~\cite{tang2024cmunext}, UltraLBM-UNet~\cite{fan2025ultralbm}, and I2U-Net~\cite{dai2024i2u}—maximize efficiency via compact convolutions or lightweight sequence designs. While highly effective, the performance gains of these methods remain strictly tied to architectural representation redesign. In contrast, rather than modifying the internal host architecture, we introduce a lightweight, plug-and-play recursive module that refines predictions externally.

Although recursive refinement has a rich history, most existing frameworks operate within the deep feature space. Consequently, they incur heavy, repeated representation-processing overhead, ranging from recurrent U-Net architectures to modern iterative pipelines anchored by external attention mechanisms \cite{alom2018recurrent,sun2026iseg}. This fundamentally differs from our setting, where the correction mechanism acts exclusively in the prediction space and continually references a fixed, single-pass anchor prediction. By decoupling refinement from deep feature manipulation, this architectural design enables highly effective correction with negligible parameter and computational overhead.

While architectural adjustments focus on strengthening the representation pathway, parallel efforts have explored improving segmentation through training-side supervision. Among these, LoMix~\cite{rahman2026lomix} provides a representative reference by dynamically regularizing and combining decoder logits; however, it remains a training-time supervision policy and does not introduce recursive inference behavior. Following this line, standard deep supervision variants—including final-logit supervision (+LL), auxiliary pathways with fixed weighting (+DS), and the fixed additive multi-scale logit aggregation introduced by MERIT (+Mutation)~\cite{rahman2024multi}—remain static once deployment begins. In parallel, heavier architecture-side refinement references such as +Harmonizing~\cite{wu2024harmonizing} pursue refinement through a cascaded dual-stage U-Net design with attention fusion, thereby modifying the host architecture itself. By contrast, our method isolates a different point in this comparison space: it operates at inference time, keeps the host backbone fixed after anchor generation, and performs lightweight recursive correction directly in prediction space.

\section{Method}
\begin{figure}[t]
\centering

\makebox[\textwidth][c]{\includegraphics[width=1.2\textwidth]{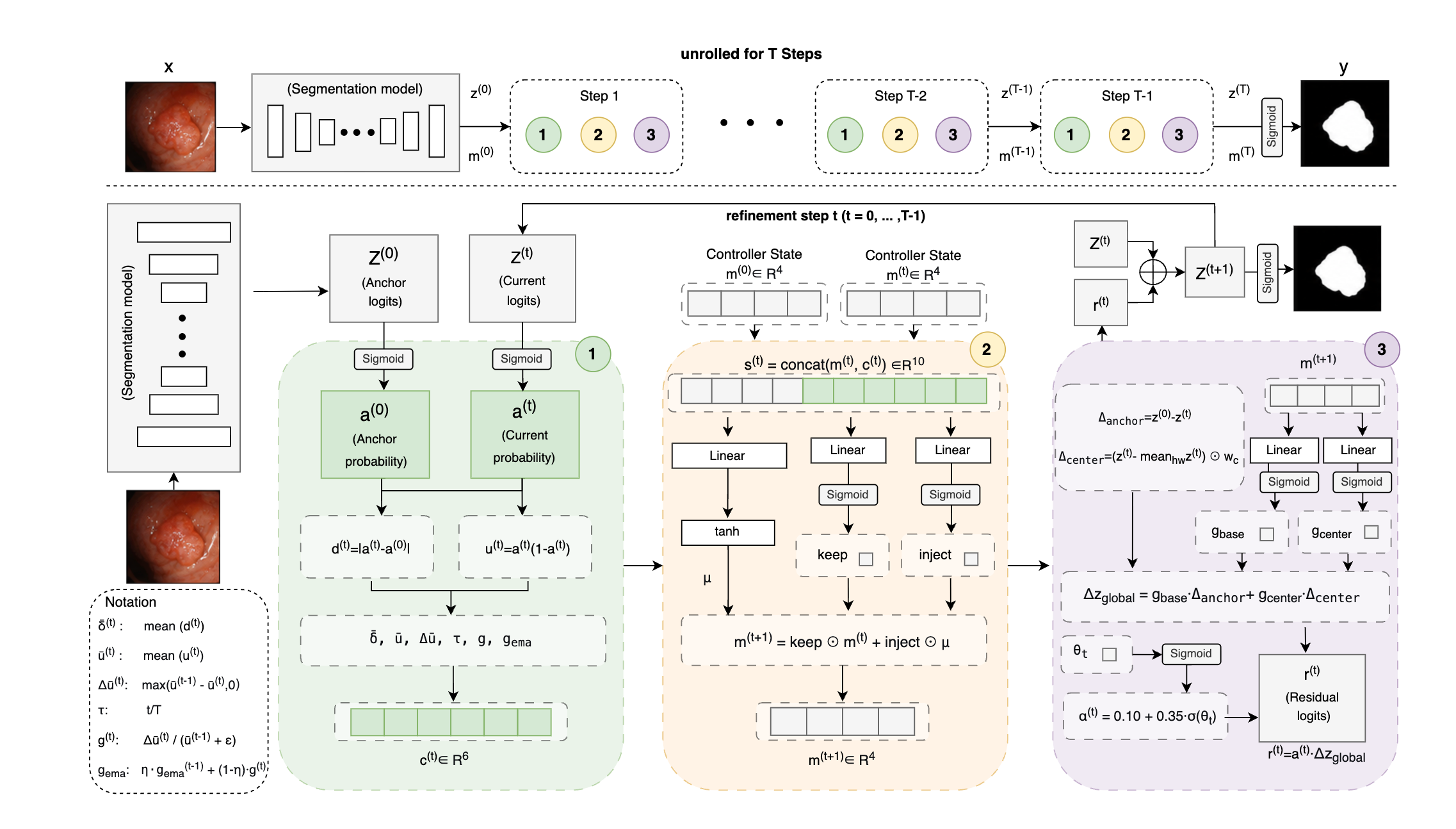}}
\caption{Overview of the gain-aware prediction-space recursive controller. The upper panel illustrates the overall framework of the proposed method, which consists of a lightweight backbone followed by $T$ iterative three-stage recursive controllers. The lower panel presents the detailed architecture and internal structure of the controller.
}
\label{fig:method_overview}
\end{figure}

In this section, we present the proposed methodology, whose overall architecture is illustrated in Fig.~\ref{fig:method_overview}. As depicted, the pipeline begins with a lightweight backbone—instantiated via a representative segmentation network—which processes an input image $x$ to generate the initial anchor logits $z^{(0)}$. To progressively refine the initial coarse logits into a precise representation, we introduce a recursive controller that operates over $T$ iterations and comprises three stages: \emph{Prediction Evidence Encoding}, \emph{Gain-Aware State Update}, and \emph{State-Guided Residual Correction}. The refined logits, $z^{(T)}$, are then passed through a Sigmoid activation function to produce the final segmentation map $y$.


\subsection{Prediction Evidence Encoding}
At this stage, the recursive controller will calculate and obtain descriptive evidences from the prediction and encode, to update and maintain the Gain-Aware State in iterations. Time step $t$, the current logits $z^{(t)}$ and the initial logits $z^{(0)}$ are first transformed into their corresponding predicted-segmentation maps, $a^{(t)}$ and $a^{(0)}$, respectively, by the Sigmoid activation function $\sigma$, shown as follows:
\begin{align}
a^{(t)} = \sigma\left(z^{(t)}\right), \quad a^{(0)} = \sigma\left(z^{(0)}\right).
\end{align}
Then we compute the discrepancy $d^{(t)}$ between $a^{(t)}$ and $a^{(0)}$, along with the uncertainty term $u^{(t)}$~\cite{kendall2017uncertainties}:
\begin{align}
d^{(t)} = \left|a^{(t)} - a^{(0)}\right|, \quad
u^{(t)} = a^{(t)}\left(1 - a^{(t)}\right).
\end{align}
Subsequently, the discrepancy map and the uncertainty map are aggregated into image-level representations via global average:
\begin{equation}
\bar d^{(t)}=\mathrm{mean}\!\left(d^{(t)}\right), \qquad 
\bar{u}^{(t)}=\mathrm{mean}\!\left(u^{(t)}\right).
\end{equation}
Additionally, the reduction in mean uncertainty between the previous and current time steps, denoted by $\Delta \bar{u}^{(t)}$, is calculated and constrained to be nonnegative, ensuring the validity of the uncertainty measure:
\begin{align}
\Delta \bar{u}^{(t)} = \max\left(\bar{u}^{(t-1)} - \bar{u}^{(t)}, 0\right).
\end{align}
Alongside $\Delta \bar{u}^{(t)}$, we define the normalized time step $\tau^{(t)}$ and the relative uncertainty reduction rate $g^{(t)}$ at step $t$ as follows:
\begin{align}
\tau^{(t)} = \frac{t}{T}, \quad g^{(t)} = \frac{\Delta \bar{u}^{(t)}}{\bar{u}^{(t-1)} + \epsilon},
\end{align}
where $\epsilon$ is a small positive constant introduced to avoid division by zero. To obtain a smoother relative uncertainty reduction rate, we introduce an exponential moving average (ema) variant, $g_{\text{ema}}^{(t)}$, defined as:
\begin{align}
g_{\text{ema}}^{(t)} &= \eta g_{\text{ema}}^{(t-1)} + (1-\eta)g^{(t)},
\end{align}
where $\eta \in [0, 1)$ denotes the smoothing momentum. Finally, we concatenate these variables into a compact state descriptor vector:
\begin{align}
c^{(t)} = \left[ \bar d^{(t)}, \bar u^{(t)}, \Delta \bar u^{(t)}, \tau^{(t)}, g^{(t)}, g_{\text{ema}}^{(t)} \right].
\label{ccc}
\end{align}
where the boundary conditions are explicitly set as $\Delta \bar u^{(0)}=0$, $g^{(0)}=0$, and $g_{\text{ema}}^{(0)}=0$. 
As indicated by equation~(\ref{ccc}), this formulation integrates error, uncertainty, and temporal dynamics between the predicted logits at time $t$ and the initial coarse logits. During $T$ iterations, the model progressively minimizes these discrepancies to converge toward the ground-truth segmentation map.

\subsection{Gain-Aware State Update}
To stabilize the direction of the correction within iterations and avoid oscillations, we introduce a memory state $m^{(t)}$ that encodes the accumulated trend of historical corrections. This state is initialized as $m^{(0)}=0$. At each iteration step $t$, we concatenate the current memory $m^{(t)}$ with the current evidence vector $c^{(t)}$, forming $s^{(t)}=[m^{(t)}; c^{(t)}]$. Subsequently, the combined vector $s^{(t)}$ is processed by three parallel linear layers to generate, respectively:


\begin{equation}
\begin{aligned}
&\mu^{(t)} = \tanh(w_i s^{(t)} + b_i), \\
&\mathrm{keep}^{(t)} = \sigma(w_j s^{(t)} + b_j), \\
&\mathrm{inject}^{(t)} = \sigma(w_k s^{(t)} + b_k).
\end{aligned}
\end{equation}
where $w_{.}$ and $b_{.}$ denote the weights and biases of the respective layers.
Specifically, the retention gate $\mathrm{keep}^{(t)}$ determines how much the historical correction trend is retained, the injection gate $\mathrm{inject}^{(t)}$ controls the integration of new evidence-conditioned information, and $\mu^{(t)}$ serves as a candidate new memory vector, derived from the concatenated vector $s^{(t)}$.
These variables are then combined to update the correction state to $m^{(t+1)}$ via a gated recurrent mechanism:
\begin{equation}
m^{(t+1)} = \mathrm{keep}^{(t)} \odot m^{(t)} + \mathrm{inject}^{(t)} \odot \mu^{(t)}.
\end{equation}
Based on the equation above, $m^{(t)}$ is recursively updated at this stage. 

\subsection{State-Guided Residual Correction}
In the third stage, the updated state $m^{(t+1)}$ is used to generate a residual correction $r^{(t)}$ that is added to the current logits $z^{(t)}$. To determine how much each correction source should contribute, two gating vectors are derived from $m^{(t+1)}$, both gates use a linear layer followed by sigmoid activation, with separate parameters:
\begin{equation}
\begin{aligned}
&g_{\mathrm{base}}^{(t)} = \sigma(w_x m^{(t+1)} + b_x), \\
&g_{\mathrm{center}}^{(t)} = \sigma(w_y m^{(t+1)} + b_y).
\end{aligned}
\end{equation}
These vectors are subsequently used to modulate and aggregate the anchor and center coordinate shifts, yielding the total global update $\Delta z_{\mathrm{global}}^{(t)}$:
\begin{equation}
\Delta z_{\mathrm{global}}^{(t)}=
g_{\mathrm{base}}^{(t)}\Delta_{anchor}
+g_{\mathrm{center}}^{(t)}\Delta_{center}\odot \text{w}_{c},
\end{equation}
where $\Delta{\mathrm{anchor}} = z^{(0)} - z^{(t)}$ represents the anchor displacement, $\Delta_{\mathrm{center}} = z^{(t)} - \bar{z}^{(t)}$ denotes the centered-logits offset, $\bar{z}^{(t)}=\mathrm{mean}_{hw}(z^{(t)})$ is the per-class spatial mean of logits over the height and width dimensions (broadcast back to the spatial grid), and $\text{w}_{c}$ is a class-wise rescaling tensor.
In addition, we introduce an empirical rescaling factor, $\alpha^{(t)}=0.10+0.35\sigma(\theta_t)$, where $\theta_t$ denotes trainable step-wise logits. Unlike a fixed per-step constant, $\alpha^{(t)}$ is adaptively learned at each iteration to regulate the update magnitude. The final logit update is then defined as
\begin{equation}
z^{(t+1)} = z^{(t)} + r^{(t)} .
\end{equation}
where $r^{(t)} = \alpha^{(t)} \cdot \Delta z_{\mathrm{global}}^{(t)}$, denotes the net residual correction applied to the predicted logits after the three processing stages.

As detailed in the preceding subsections, the operational workflow of our proposed framework is formalized in Algorithm~\ref{alg:controller}. For model optimization, the model is trained using a hybrid objective function defined as:
$$\mathcal{L}_{\mathrm{total}} = 0.6\,\mathcal{L}_{\mathrm{Dice}} + 0.4\,\mathcal{L}_{\mathrm{Focal}}$$
where $\mathcal{L}_{\mathrm{Dice}}$ denotes the standard Dice loss as formulated in \cite{wu2024harmonizing}, and the focal component $\mathcal{L}_{\mathrm{Focal}}$ follows the well-established dense-prediction framework introduced by \cite{lin2017focal}.

\begin{algorithm}[t]
\caption{Recursive Prediction-Space Correction}
\label{alg:controller}
\begin{algorithmic}[1]
\Require Image $x$, backbone $f$, loop budget $T$
\Ensure Final probability map $y$
\State $z^{(0)} \gets f(x)$
\State $a^{(0)} \gets \sigma(z^{(0)})$
\State $m^{(0)} \gets \mathbf{0}$
\State $\Delta\bar u^{(0)} \gets 0$, \quad $g^{(0)} \gets 0$, \quad $g_{\mathrm{ema}}^{(0)} \gets 0$
\For{$t = 0$ to $T-1$}
    \State $a^{(t)} \gets \sigma(z^{(t)})$
    \State $d^{(t)} \gets |a^{(t)}-a^{(0)}|$, \quad $u^{(t)} \gets a^{(t)}(1-a^{(t)})$
    \State $\bar d^{(t)} \gets \mathrm{mean}(d^{(t)})$, \quad $\bar u^{(t)} \gets \mathrm{mean}(u^{(t)})$
    \If{$t > 0$}
        \State $\Delta\bar u^{(t)} \gets \max(\bar u^{(t-1)}-\bar u^{(t)},0)$
        \State $g^{(t)} \gets \Delta\bar u^{(t)}/(\bar u^{(t-1)}+\epsilon)$
        \State $g_{\mathrm{ema}}^{(t)} \gets \eta\,g_{\mathrm{ema}}^{(t-1)} + (1-\eta)\,g^{(t)}$
    \EndIf
    \State $\tau^{(t)} \gets t/T$, \quad $c^{(t)} \gets [\bar d^{(t)},\bar u^{(t)},\Delta\bar u^{(t)},\tau^{(t)},g^{(t)},g_{\mathrm{ema}}^{(t)}]$
    \State $s^{(t)} \gets [m^{(t)};c^{(t)}]$
    \State compute $\mu^{(t)}$, $\mathrm{keep}^{(t)}$, and $\mathrm{inject}^{(t)}$ from $s^{(t)}$
    \State $m^{(t+1)} \gets \mathrm{keep}^{(t)} \odot m^{(t)} + \mathrm{inject}^{(t)} \odot \mu^{(t)}$
    \State compute $g_{\mathrm{base}}^{(t)}$ and $g_{\mathrm{center}}^{(t)}$ from $m^{(t+1)}$
    \State $\Delta z_{\mathrm{global}}^{(t)} \gets g_{\mathrm{base}}^{(t)}(z^{(0)}-z^{(t)}) + g_{\mathrm{center}}^{(t)}(z^{(t)}-\bar z^{(t)})\odot w_{\mathrm{c}}$
    \State $r^{(t)} \gets \alpha^{(t)} \cdot \Delta z_{\mathrm{global}}^{(t)}$
    \State $z^{(t+1)} \gets z^{(t)} + r^{(t)}$
\EndFor
\State $y \gets \sigma(z^{(T)})$
\State \Return $y$
\end{algorithmic}
\end{algorithm}

\section{Experiments}
\subsection{Experimental Setup}
Unless otherwise specified, all models are trained on the Kvasir-SEG~\cite{jha2019kvasir} dataset and evaluated under a unified cross-dataset protocol, where generalization capability is assessed through zero-shot transfer on CVC-ClinicDB~\cite{bernal2015wm}, CVC-ColonDB~\cite{tajbakhsh2015automated}, and PolypGen~\cite{ali2023multi}. All models are trained for 200 epochs using the AdamW optimizer with cosine learning rate decay, a batch size of 16, and an input resolution of $352 \times 352$. Model selection is based exclusively on the validation Dice score. To ensure fair comparisons, all baseline variants adopt the same primary supervision framework. Unless explicitly stated otherwise, the recursive controller is executed with a fixed computational budget of $T=5$ iterations throughout all experiments. For quantitative evaluation, both the Dice coefficient and Intersection-over-Union (IoU) are reported following standard lightweight segmentation conventions, while the Dice score alone is used for checkpoint selection. In addition, parameter count and profiled GMACs are reported as static measures of model complexity, whereas peak GPU memory consumption is used to evaluate practical runtime deployment efficiency.

To ensure clarity across subsequent empirical evaluations, we formally define the comparative configurations instantiated on each host backbone:
\begin{itemize}
\item  +LL specifies standard supervision applied exclusively to the final decoder logits.
\item +DS denotes deep auxiliary supervision across intermediate decoder outputs using static, fixed weighting.
\item +Mutation replicates the fixed, additive multi-scale logit aggregation strategy introduced by MERIT~\cite{rahman2024multi}.
\item +LoMix represents a training-side policy utilizing learnable weighted logit mixing~\cite{rahman2026lomix}.
\item +Harmonizing implements a heavier architecture-side refinement reference featuring a cascaded dual-stage attention-fusion design adapted from H-Unets~\cite{wu2024harmonizing}.
\end{itemize}
Crucially, these configurations serve as external comparison protocols applied to identical host backbones; they do not represent native modules inherent to the underlying host architectures.


\subsection{Experimental Results}
\subsubsection{Main Results}
\begin{figure*}[!t]
\centering
\includegraphics[width=\textwidth]{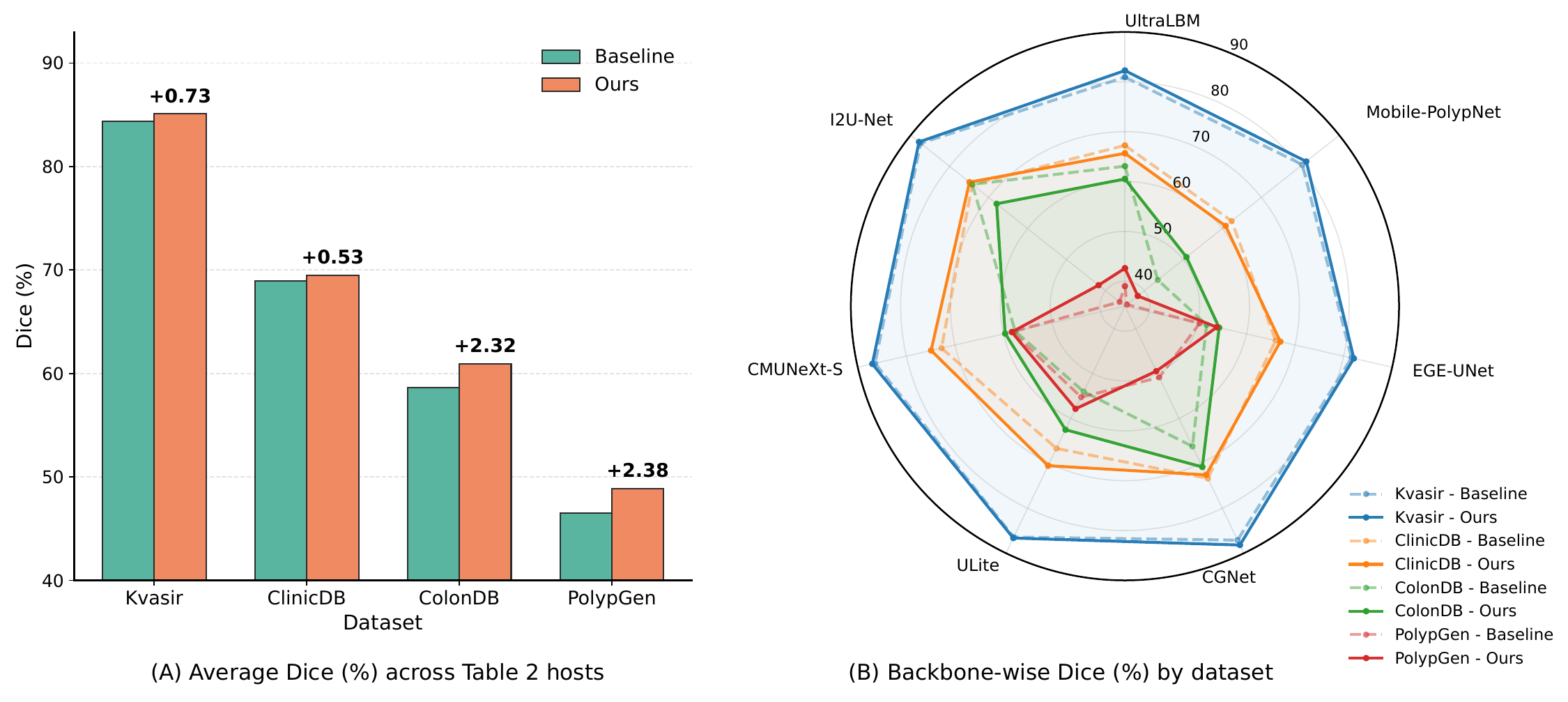}
\caption{Visualization of Table~\ref{tab:main_results}. Panel~(A) summarizes average Dice gains, and Panel~(B) reorganizes the same evidence by backbone and dataset to highlight stable source-domain gains and backbone-dependent transfer behavior.}
\label{fig:main_results_visualization}
\end{figure*}

\begin{table*}[!t]
\caption{Main multi-dataset results under the unified Kvasir-trained protocol. `Kvasir$\uparrow$` denotes source-domain IoU/Dice on Kvasir-SEG \cite{jha2019kvasir}, while `K$\rightarrow$ClinicDB$\uparrow$`, `K$\rightarrow$ColonDB$\uparrow$`, and `K$\rightarrow$PolypGen$\uparrow$` denote transfer IoU/Dice on CVC-ClinicDB \cite{bernal2015wm}, CVC-ColonDB \cite{tajbakhsh2015automated}, and PolypGen \cite{ali2023multi}. Bold marks the better result within each baseline/controller pair.}
\label{tab:main_results}
\centering
\scriptsize
\setlength{\tabcolsep}{4.2pt}
\resizebox{\textwidth}{!}{
\begin{tabular}{llcccccc}
\toprule
Backbone & Variant & \makecell{Kvasir$\uparrow$\\IoU / Dice} & \makecell{K$\rightarrow$ClinicDB$\uparrow$\\IoU / Dice} & \makecell{K$\rightarrow$ColonDB$\uparrow$\\IoU / Dice} & \makecell{K$\rightarrow$PolypGen$\uparrow$\\IoU / Dice} & \makecell{Params(M)} & \makecell{GMACs} \\
\midrule
UltraLBM-UNet & Baseline & 0.6798 / 0.8095 & \textbf{0.5067 / 0.6726} & \textbf{0.4608 / 0.6309} & 0.2425 / 0.3903 & 0.0344 & 24.6704 \\
 & Proposed Controller (Ours) & \textbf{0.6986 / 0.8227} & 0.4889 / 0.6567 & 0.4339 / 0.6052 & \textbf{0.2707 / 0.4261} & 0.0346& 24.6704 \\
\midrule
Mobile-PolypNet & Baseline & 0.6741 / 0.8053 & \textbf{0.4534 / 0.6240} & 0.2774 / 0.4343 & 0.2160 / 0.3552 & 0.1492 & 1.5836 \\
 & Proposed Controller (Ours) & \textbf{0.6885 / 0.8155} & 0.4370 / 0.6083 & \textbf{0.3404 / 0.5079} & \textbf{0.2369 / 0.3830} & 0.1494& 1.5836 \\
\midrule
EGE-UNet & Baseline & 0.6893 / 0.8163 & 0.4927 / 0.6602 & 0.3496 / 0.5181 & 0.3368 / 0.5038 & 0.2205 & 2.3193 \\
 & Proposed Controller (Ours) & \textbf{0.6966 / 0.8212} & \textbf{0.5035 / 0.6698} & \textbf{0.3735 / 0.5439} & \textbf{0.3691 / 0.5392} & 0.2207 & 2.3193 \\
\midrule
CGNet & Baseline & 0.7713 / 0.8709 & \textbf{0.5791 / 0.7335} & 0.4949 / 0.6621 & \textbf{0.3410 / 0.5085} & 0.4920 & 3.3600 \\
 & Proposed Controller (Ours) & \textbf{0.7886 / 0.8818} & 0.5694 / 0.7256 & \textbf{0.5480 / 0.7080} & 0.3287 / 0.4947 & 0.4922 & 3.3600 \\
\midrule
ULite & Baseline & 0.7607 / 0.8641 & 0.5001 / 0.6667 & 0.3703 / 0.5405 & 0.3816 / 0.5525 & 0.8784 & 1.4314 \\
 & Proposed Controller (Ours) & \textbf{0.7641 / 0.8663} & \textbf{0.5447 / 0.7052} & \textbf{0.4546 / 0.6250} & \textbf{0.4071 / 0.5786} & 0.8786 & 1.4314 \\
\midrule
CMUNeXt-S & Baseline & 0.7613 / 0.8645 & 0.5718 / 0.7276 & 0.4044 / 0.5759 & 0.4101 / 0.5817 & 1.8333 & 13.0744 \\
 & Proposed Controller (Ours) & \textbf{0.7699 / 0.8700} & \textbf{0.5987 / 0.7490} & \textbf{0.4251 / 0.5966} & \textbf{0.4113 / 0.5829} & 1.8335 & 13.0744 \\
\midrule
I2U-Net & Baseline & 0.7767 / 0.8743 & 0.5896 / 0.7418 & \textbf{0.5895 / 0.7418} & 0.2223 / 0.3637 & 7.0318 & 6.7663 \\
 & Proposed Controller (Ours) & \textbf{0.7833 / 0.8785} & \textbf{0.5987 / 0.7490} & 0.5145 / 0.6794 & \textbf{0.2639 / 0.4176} & 7.0320 & 6.7663 \\
\bottomrule
\end{tabular}
\vspace{1pt}
}
\end{table*}
Table~\ref{tab:main_results} reports the primary evaluation results under the unified Kvasir-trained protocol across seven lightweight backbones, including UltraLBM-UNet~\cite{fan2025ultralbm}, Mobile-PolypNet~\cite{karmakar2022mobile}, EGE-UNet~\cite{ruan2023ege}, CGNet~\cite{wu2020cgnet}, ULite~\cite{dinh20231m}, CMUNeXt-S~\cite{tang2024cmunext}, and I2U-Net~\cite{dai2024i2u}. Each backbone is reported as a paired baseline/controller row, allowing the gain of the proposed controller to be read under fixed hosts. Fig.~\ref{fig:main_results_visualization} summarizes the same evidence from two complementary views. Panel~(A) reports the average Dice improvement across the seven backbones for each dataset, showing uniformly positive source-domain gains on Kvasir-SEG and larger average gains on the more challenging transfer sets ColonDB and PolypGen. Panel~(B) reorganizes the results by backbone and dataset, making the host-dependent transfer pattern easier to inspect. Overall, source-domain gains on Kvasir-SEG remain positive across all seven hosts, while transfer gains are selective rather than universal. Notably, EGE-UNet, ULite, and CMUNeXt-S achieve positive improvements on all three transfer datasets, whereas other hosts exhibit more uneven transfer behavior. These results support a focused conclusion: the proposed controller consistently improves same-host source-domain segmentation and can deliver meaningful transfer gains on selected hosts without changing the underlying backbone.

\subsubsection{Cross-Backbone Kvasir Comparison}

\begin{table*}[!t]
\caption{Cross-backbone Kvasir comparison in row-block form. `+LL` is the per-backbone anchor; `+DS`, `+Mutation`, and `+LoMix` follow the training-side comparison line of MERIT/LoMix \cite{rahman2024multi,rahman2026lomix}; `+Harmonizing` is the heavier cascaded attention-fusion reference adapted from H-Unets \cite{wu2024harmonizing}. Bold and underline indicate the best and second-best variants within each backbone block.}
\label{tab:cross_backbone_plugins}
\centering
\scriptsize
\setlength{\tabcolsep}{5.0pt}
\begin{tabular}{llcc}
\toprule
Backbone & Variant & IoU / Dice & $\Delta$IoU / $\Delta$Dice \\
\midrule
\multirow{6}{*}{Mobile-PolypNet} & +LL (Baseline) & 0.6741 / 0.8053 & -- / -- \\
 & +DS & 0.6774 / 0.8077 & +0.33\% / +0.24\% \\
 & +Mutation & 0.6678 / 0.8008 & -0.63\% / -0.45\% \\
 & +LoMix & 0.6653 / 0.7990 & -0.88\% / -0.63\% \\
 & +Harmonizing & \underline{0.6858 / 0.8136} & \underline{+1.17\% / +0.83\%} \\
 & +Proposed Controller (Ours) & \textbf{0.6885 / 0.8155} & \textbf{+1.44\% / +1.02\%} \\
\midrule
\multirow{6}{*}{UltraLBM-UNet} & +LL (Baseline) & 0.6798 / 0.8095 & -- / -- \\
 & +DS & 0.6908 / 0.8171 & +1.10\% / +0.76\% \\
 & +Mutation & 0.6685 / 0.8013 & -1.13\% / -0.82\% \\
 & +LoMix & 0.6681 / 0.8010 & -1.17\% / -0.85\% \\
 & +Harmonizing & \underline{0.6933 / 0.8189} & \underline{+1.35\% / +0.94\%} \\
 & +Proposed Controller (Ours) & \textbf{0.6988 / 0.8227} & \textbf{+1.90\% / +1.32\%} \\
\midrule
\multirow{6}{*}{CMUNeXt-S} & +LL (Baseline) & 0.7613 / 0.8645 & -- / -- \\
 & +DS & 0.7660 / 0.8675 & +0.47\% / +0.30\% \\
 & +Mutation & 0.7624 / 0.8652 & +0.11\% / +0.07\% \\
 & +LoMix & \textbf{0.7803 / 0.8766} & \textbf{+1.90\% / +1.21\%} \\
 & +Harmonizing & 0.7595 / 0.8633 & -0.18\% / -0.12\% \\
 & +Proposed Controller (Ours) & \underline{0.7699 / 0.8700} & \underline{+0.86\% / +0.55\%} \\
\midrule
\multirow{6}{*}{EGE-UNet} & +LL (Baseline) & 0.6893 / 0.8163 & -- / -- \\
 & +DS & 0.6939 / 0.8193 & +0.46\% / +0.30\% \\
 & +Mutation & 0.6951 / 0.8201 & +0.58\% / +0.38\% \\
 & +LoMix & 0.6902 / 0.8167 & +0.09\% / +0.04\% \\
 & +Harmonizing & \underline{0.6955 / 0.8204} & \underline{+0.62\% / +0.41\%} \\
 & +Proposed Controller (Ours) & \textbf{0.6966 / 0.8212} & \textbf{+0.73\% / +0.49\%} \\
\bottomrule
\end{tabular}
\end{table*}

\begin{figure*}[!t]
\centering
\includegraphics[width=\textwidth]{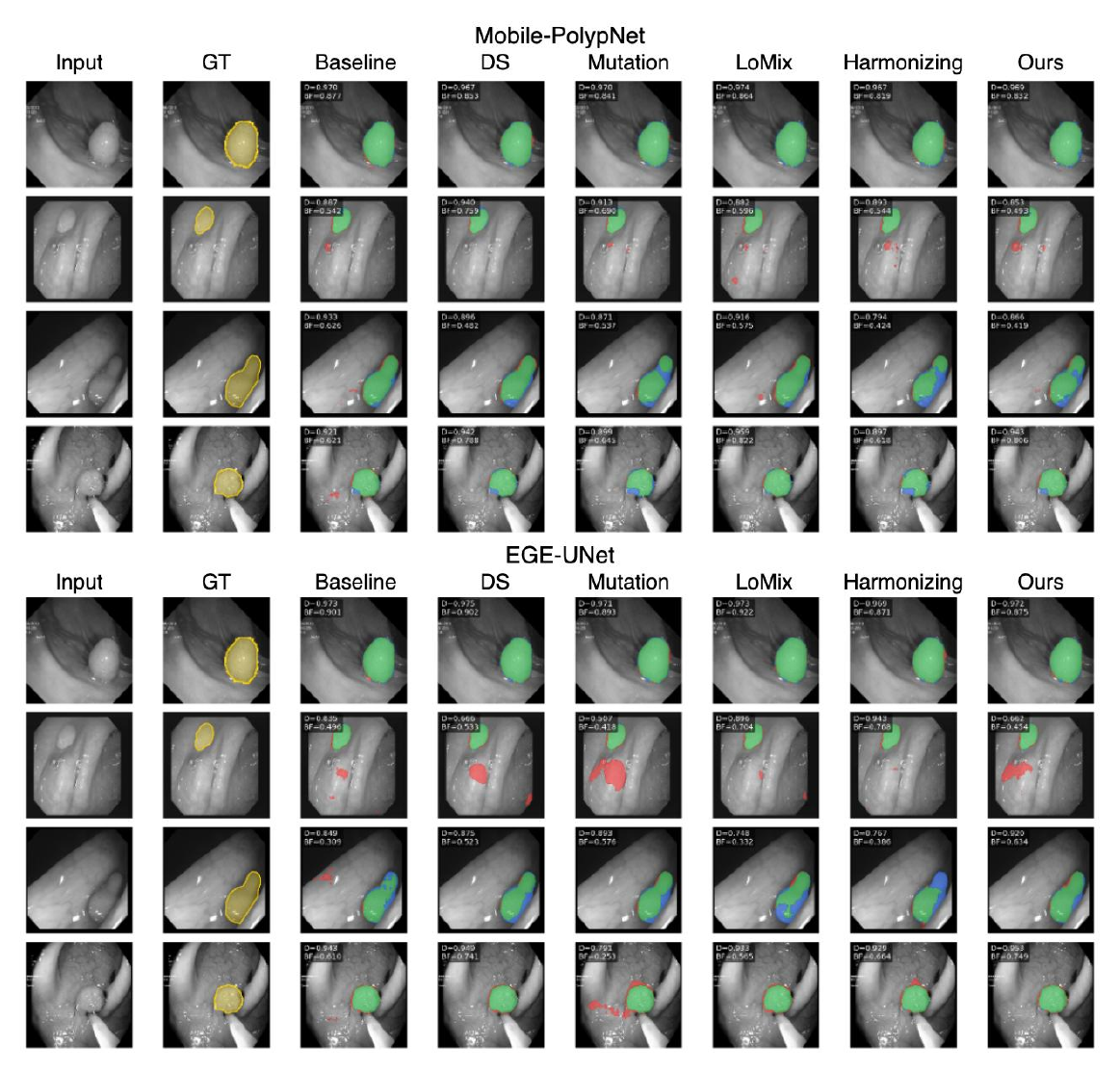}
\caption{Qualitative comparison on Mobile-PolypNet \cite{karmakar2022mobile} and EGE-UNet \cite{ruan2023ege} using one representative case from each dataset: Kvasir-SEG \cite{jha2019kvasir}, CVC-ClinicDB \cite{bernal2015wm}, CVC-ColonDB \cite{tajbakhsh2015automated}, and PolypGen \cite{ali2023multi}. Columns report Input, GT, Baseline, `+DS`, `+Mutation`, `+LoMix`, `+Harmonizing`, and Ours. Green, red, and blue denote correct coverage, false positives, and false negatives, respectively.}
\label{fig:qualitative}
\end{figure*}

Table~\ref{tab:cross_backbone_plugins} evaluates performance under the same Kvasir protocol by comparing our plug-and-play controller against training-side policies and a heavier structural refiner. To reflect distinct architectural scales and design styles, we report results on four representative backbones: Mobile-PolypNet, UltraLBM-UNet, CMUNeXt-S, and EGE-UNet.

Under this setup, +LL, +DS, +Mutation, and +LoMix represent training-side supervision or logits-mixing protocols, whereas +Harmonizing serves as a heavier architecture-side refinement reference. By contrast, Ours is the only inference-time prediction-space controller in this comparison. The quantitative results in Table~\ref{tab:cross_backbone_plugins} show that our method remains positive across all four tested backbones, while the stronger training-side baselines are more host-dependent. In particular, +Mutation and +LoMix reduce performance on Mobile-PolypNet and UltraLBM-UNet, with the largest IoU drop reaching -1.17\%. Ours achieves the best IoU in the Mobile-PolypNet, UltraLBM-UNet, and EGE-UNet blocks, and remains competitive on CMUNeXt-S. It also exceeds the heavier +Harmonizing reference in all four backbone blocks. These results position the proposed controller as a lightweight and stable alternative within this comparison space, rather than as a universally dominant method across all training-side policies.

Fig.~\ref{fig:qualitative} visually complements Table~\ref{tab:cross_backbone_plugins} at the boundary level, focusing on representative cases across Kvasir-SEG, CVC-ClinicDB, CVC-ColonDB, and PolypGen. Within this tightly contested setting, all methods generally localize the lesion, shifting the true evaluation to finer, edge-level error behavior. The visual evidence clearly distinguishes our approach: our controller consistently suppresses scattered red false-positive regions around the lesion periphery while simultaneously recovering thin blue false-negative boundary fragments. This superior refinement is especially visible on the cross-dataset transfer rows, where baseline and training-side variants suffer from obvious edge leakage or missed contours, whereas our model produces clean, tightly-bounded, and anatomically precise segmentations.

\subsubsection{Component Ablation}
\begin{table*}[!t]
\caption{Component ablation on CMUNeXt-S \cite{tang2024cmunext} and CGNet \cite{wu2020cgnet}. All controller rows use the default fixed-step budget $T=5$. `w/o strengthened global correction` weakens the residual branch, `w/o gain-aware state` removes the state update, and `Full (Ours)` restores the complete controller. Bold marks the best setting within each backbone block.}
\label{tab:ablation}
\centering
\scriptsize
\setlength{\tabcolsep}{6pt}
\begin{tabular}{llcccc}
\toprule
Backbone & Setting & IoU$\uparrow$ & Dice$\uparrow$ & $\Delta$IoU & $\Delta$Dice \\
\midrule
CMUNeXt-S & Backbone only & 0.7613 & 0.8645 & -- & -- \\
 & w/o strengthened global correction & 0.7665 & 0.8678 & +0.52\% & +0.33\% \\
 & w/o gain-aware state & 0.7690 & 0.8694 & +0.77\% & +0.49\% \\
 & Full (Ours) & \textbf{0.7699} & \textbf{0.8700} & \textbf{+0.86\%} & \textbf{+0.55\%} \\
\midrule
CGNet & Backbone only & 0.7713 & 0.8709 & -- & -- \\
 & w/o strengthened global correction & 0.7794 & 0.8760 & +0.81\% & +0.51\% \\
 & w/o gain-aware state & 0.7674 & 0.8684 & -0.39\% & -0.25\% \\
 & Full (Ours) & \textbf{0.7886} & \textbf{0.8818} & \textbf{+1.73\%} & \textbf{+1.09\%} \\
\bottomrule
\end{tabular}
\end{table*}
Table~\ref{tab:ablation} presents a component ablation study isolating the individual contributions of the gain-aware state and strengthened global correction branches using CMUNeXt-S and CGNet \cite{tang2024cmunext,wu2020cgnet} as representatives of contrasting baseline correction sensitivities. The empirical results reveal distinct functional roles for each module across the two architectures. For CMUNeXt-S, omitting either component degrades performance relative to the complete configuration, though both variants still outperform the standalone backbone; the integrated controller (Full) yields the top performance, with gains of $0.86\%$ $\Delta$IoU and $0.55\%$ $\Delta$Dice. Conversely, on CGNet, the gain-aware state emerges as the primary driver of stability. Removing the state update drops performance below the backbone-only baseline ($-0.39\%$ $\Delta$IoU), whereas weakening the global correction branch while preserving the state still maintains a net positive gain ($+0.81\%$ $\Delta$IoU). This behavioral divergence suggests that for host architectures less inherently stable under recursive estimation, the quality of state modulation is more critical to overall performance than raw residual correction strength.

\subsubsection{Efficiency}
\begin{table*}[!t]
\caption{Added-cost comparison against the baseline (`+LL`) for the two deployed refinement variants reported in the paper: `+Harmonizing` \cite{wu2024harmonizing} and Ours. `$\Delta$Params` and `$\Delta$GMACs` are static architectural increments profiled with THOP; for Ours, `$\Delta$GMACs=0.0000` indicates a negligible increment under the adopted profiling precision rather than literal zero runtime computation. `$\Delta$Peak Mem.` is the runtime deployment indicator. Smaller added cost is better in all three columns.}


\label{tab:added_cost}
\centering
\scriptsize
\setlength{\tabcolsep}{5.5pt}
\begin{tabular}{llccc}
\toprule
Backbone & Variant & \makecell{$\Delta$Params\\(params)} & $\Delta$GMACs & \makecell{$\Delta$Peak Mem.\\(MB)} \\
\midrule
\multirow{2}{*}{Mobile-PolypNet} & +Harmonizing & +170378 & +1.9036 & +49.0 \\
 & +Proposed Controller (Ours) & +189 & +0.0000 & +37.7 \\
\midrule
\multirow{2}{*}{UltraLBM-UNet} & +Harmonizing & +51081 & +0.1737 & +6.9 \\
 & +Proposed Controller (Ours) & +189 & +0.0000 & +4.9 \\
\midrule
\multirow{2}{*}{CMUNeXt-S} & +Harmonizing & +2095246 & +14.8923 & +65.2 \\
 & +Proposed Controller (Ours) & +189 & +0.0000 & +37.7 \\
\midrule
\multirow{2}{*}{EGE-UNet} & +Harmonizing & +368269 & +3.3762 & +46.3 \\
 & +Proposed Controller (Ours) & +189 & +0.0000 & +37.0 \\
\bottomrule
\end{tabular}
\end{table*}
Table~\ref{tab:added_cost} quantifies the deployment overhead introduced by +Harmonizing and our proposed controller relative to the baseline (+LL) across the four host architectures. The empirical comparison reveals a stark contrast in efficiency: while +Harmonizing imposes substantial static overhead—requiring up to $2,095,246$ additional parameters and $14.8923$ GMACs on CMUNeXt-S—our controller maintains a negligible, constant footprint of just $189$ parameters and a THOP-profiled GMAC increment reported as $0.0000$ at the adopted precision. Furthermore, our approach consistently demands fewer runtime resources, yielding lower peak memory increments across the board (e.g., $37.7$ MB versus $65.2$ MB on CMUNeXt-S). Serving as the deployment-side counterpart to Table~\ref{tab:cross_backbone_plugins}, these metrics demonstrate that competitive refinement accuracy can be achieved without inheriting the prohibitive structural and computational costs of an auxiliary refinement path.


\subsubsection{Loop Budget}
\begin{table}[!t]
\caption{Fixed-step sensitivity of the proposed controller on CMUNeXt-S \cite{tang2024cmunext} and CGNet \cite{wu2020cgnet}. Higher IoU/Dice is better. The best loop depth is backbone-dependent. Bold marks the better loop depth within each host column.}
\label{tab:loop_budget}
\centering
\scriptsize
\setlength{\tabcolsep}{7pt}
\begin{tabular}{lcc}
\toprule
Loop Budget $T$ & \makecell{CMUNeXt-S\\IoU / Dice} & \makecell{CGNet\\IoU / Dice} \\
\midrule
$T=1$ & 0.7613 / 0.8645 & 0.7713 / 0.8709 \\
$T=3$ & 0.7649 / 0.8673 & 0.7757 / 0.8737 \\
$T=5$ & 0.7699 / 0.8700& \textbf{0.7886 / 0.8818} \\
$T=7$ & \textbf{0.7815 / 0.8775} & 0.7838 / 0.8788 \\
\bottomrule
\end{tabular}
\end{table}
Table~\ref{tab:loop_budget} evaluates the sensitivity of the proposed controller to the fixed-step loop budget $T$ across CMUNeXt-S and CGNet \cite{tang2024cmunext,wu2020cgnet}, demonstrating that recursive depth operates as a backbone-dependent trade-off rather than a monotonic scaling law. For CMUNeXt-S, performance scales continuously with loop depth, achieving its peak metrics at $T=7$ ($0.7815$ IoU / $0.8775$ Dice). Conversely, CGNet exhibits a clear saturation threshold, peaking at $T=5$ ($0.7886$ IoU / $0.8818$ Dice) before experiencing performance degradation when extended to $T=7$ ($0.7838$ IoU / $0.8788$ Dice). This divergence confirms that increasing recursive iterations does not yield universal utility; instead, loop depth must be selectively calibrated to the architectural tolerance and error-correction characteristics of each backbone.



\section{Conclusion}
In this work, we introduced a lightweight polyp segmentation framework that pairs a flexible backbone with a gain-aware, prediction-space recursive controller designed for high efficiency. Rather than expanding the representation pathway, our method freezes the backbone after generating initial anchor logits. It then routes subsequent corrections through compact discrepancy and uncertainty evidence, gated recurrent mechanism, and additive residual updates under an explicit loop budget.

The empirical evidence throughout this study consistently demonstrates that a well-defined prediction-space correction framework can deliver reliable source-domain gains, competitive performance against training-side and heavy structural refinement baselines, and a bounded deployment footprint. Consequently, our findings offer a pragmatic architectural takeaway: lightweight, inference-time correction is highly effective when formalized as an explicit prediction-space controller, bypassing the need for a costly expansion of the deep representation path.

A current limitation of this scope is that cross-domain transfer gains remain sensitive to specific architectures and boundary shifts. Extending this controller paradigm to more challenging cross-domain scenarios by leveraging richer error evidence forms a natural and compelling direction for future work.

\bibliographystyle{unsrt}
\bibliography{paper_refs_manual_v1}

@inproceedings{dinh20231m,
  title={1M parameters are enough? A lightweight CNN-based model for medical image segmentation},
  author={Dinh, Binh-Duong and Nguyen, Thanh-Thu and Tran, Thi-Thao and Pham, Van-Truong},
  booktitle={2023 Asia Pacific Signal and Information Processing Association Annual Summit and Conference (APSIPA ASC)},
  pages={1279--1284},
  year={2023},
  organization={IEEE}
}

@article{tesema2026lgps,
  title={Lgps: A lightweight gan-based approach for polyp segmentation in colonoscopy images},
  author={Tesema, Fiseha Berhanu and Guerra-Manzanares, Alejandro and Cui, Tianxiang and Zhang, Qian and Solomon, Moses M and He, Xiangjian},
  journal={Biomedical Signal Processing and Control},
  volume={118},
  pages={109777},
  year={2026},
  publisher={Elsevier}
}

@article{ramos2026multi,
  title={Multi-encoder ConvNeXt network with smooth attentional feature fusion for multispectral semantic segmentation},
  author={Ramos, Leo Thomas and Sappa, Angel D},
  journal={Neurocomputing},
  pages={133533},
  year={2026},
  publisher={Elsevier}
}

@article{zami2026ultralightcps,
  title={UltraLightCPS: An Ultra-Lightweight Colon Polyp Segmentation model using an adaptive compressed multi-scale semantic fusion mechanism},
  author={Zami, Abduz and Sobhan, Shadman and Ahmed, Mohiuddin and Uddin, Md Palash},
  journal={Biomedical Signal Processing and Control},
  volume={120},
  pages={110162},
  year={2026},
  publisher={Elsevier}
}

@article{dai2024i2u,
  title={I2u-net: A dual-path u-net with rich information interaction for medical image segmentation},
  author={Dai, Duwei and Dong, Caixia and Yan, Qingsen and Sun, Yongheng and Zhang, Chunyan and Li, Zongfang and Xu, Songhua},
  journal={Medical Image Analysis},
  volume={97},
  pages={103241},
  year={2024},
  publisher={Elsevier}
}

@article{fan2025ultralbm,
  title={UltraLBM-UNet: Ultralight Bidirectional Mamba-based Model for Skin Lesion Segmentation},
  author={Fan, Linxuan and Jiang, Juntao and Liu, Weixuan and Xue, Zhucun and Lv, Jiajun and Zhang, Jiangning and Liu, Yong},
  journal={arXiv preprint arXiv:2512.21584},
  year={2025}
}

@inproceedings{ruan2023ege,
  title={Ege-unet: an efficient group enhanced unet for skin lesion segmentation},
  author={Ruan, Jiacheng and Xie, Mingye and Gao, Jingsheng and Liu, Ting and Fu, Yuzhuo},
  booktitle={International conference on medical image computing and computer-assisted intervention},
  pages={481--490},
  year={2023},
  organization={Springer}
}

@article{wu2020cgnet,
  title={CGNet: A light-weight context guided network for semantic segmentation},
  author={Wu, Tianyi and Tang, Sheng and Zhang, Rui and Cao, Juan and Zhang, Yongdong},
  journal={IEEE Transactions on Image Processing},
  volume={30},
  pages={1169--1179},
  year={2020},
  publisher={IEEE}
}

@article{karmakar2022mobile,
  title={Mobile-PolypNet: Lightweight colon polyp segmentation network for low-resource settings},
  author={Karmakar, Ranit and Nooshabadi, Saeid},
  journal={Journal of imaging},
  volume={8},
  number={6},
  pages={169},
  year={2022},
  publisher={MDPI}
}

@inproceedings{tang2024cmunext,
  title={Cmunext: An efficient medical image segmentation network based on large kernel and skip fusion},
  author={Tang, Fenghe and Ding, Jianrui and Quan, Quan and Wang, Lingtao and Ning, Chunping and Zhou, S Kevin},
  booktitle={2024 IEEE International Symposium on Biomedical Imaging (ISBI)},
  pages={1--5},
  year={2024},
  organization={IEEE}
}

@inproceedings{rahman2024multi,
  title={Multi-scale hierarchical vision transformer with cascaded attention decoding for medical image segmentation},
  author={Rahman, Md Mostafijur and Marculescu, Radu},
  booktitle={Medical Imaging with Deep Learning},
  pages={1526--1544},
  year={2024},
  organization={PMLR}
}

@article{zhao2026dgfi,
  title={DGFI-Net: dual-branch guided feature interaction network for brain tumor segmentation},
  author={Zhao, Longyun and Jiang, Xiaoliang and Zhang, Qile},
  journal={Frontiers in Physiology},
  volume={17},
  pages={1817401},
  year={2026},
  publisher={Frontiers Media SA}
}

@article{rahman2026lomix,
  title={LoMix: Learnable Weighted Multi-Scale Logits Mixing for Medical Image Segmentation},
  author={Rahman, Md Mostafijur and Marculescu, Radu},
  journal={Advances in Neural Information Processing Systems},
  volume={38},
  pages={65778--65809},
  year={2026}
}

@article{wu2024harmonizing,
  title={Harmonizing unets: Attention fusion module in cascaded-unets for low-quality oct image fluid segmentation},
  author={Wu, Zhuoyu and Wu, Qinchen and Fang, Wenqi and Ou, Wenhui and Wang, Quanjun and Zhang, Linde and Chen, Chao and Wang, Zheng and Li, Heshan},
  journal={Computers in Biology and Medicine},
  volume={183},
  pages={109223},
  year={2024},
  publisher={Elsevier}
}

@inproceedings{jha2019kvasir,
  title={Kvasir-seg: A segmented polyp dataset},
  author={Jha, Debesh and Smedsrud, Pia H and Riegler, Michael A and Halvorsen, P{\aa}l and De Lange, Thomas and Johansen, Dag and Johansen, H{\aa}vard D},
  booktitle={International conference on multimedia modeling},
  pages={451--462},
  year={2019},
  organization={Springer}
}

@article{bernal2015wm,
  title={WM-DOVA maps for accurate polyp highlighting in colonoscopy: Validation vs. saliency maps from physicians},
  author={Bernal, Jorge and S{\'a}nchez, F Javier and Fern{\'a}ndez-Esparrach, Gloria and Gil, Debora and Rodr{\'\i}guez, Cristina and Vilari{\~n}o, Fernando},
  journal={Computerized medical imaging and graphics},
  volume={43},
  pages={99--111},
  year={2015},
  publisher={Elsevier}
}

@article{tajbakhsh2015automated,
  title={Automated polyp detection in colonoscopy videos using shape and context information},
  author={Tajbakhsh, Nima and Gurudu, Suryakanth R and Liang, Jianming},
  journal={IEEE transactions on medical imaging},
  volume={35},
  number={2},
  pages={630--644},
  year={2015},
  publisher={IEEE}
}

@article{ali2023multi,
  title={A multi-centre polyp detection and segmentation dataset for generalisability assessment},
  author={Ali, Sharib and Jha, Debesh and Ghatwary, Noha and Realdon, Stefano and Cannizzaro, Renato and Salem, Osama E and Lamarque, Dominique and Daul, Christian and Riegler, Michael A and Anonsen, Kim V and others},
  journal={Scientific Data},
  volume={10},
  number={1},
  pages={75},
  year={2023},
  publisher={Nature Publishing Group UK London}
}

@article{yue2024boundary,
  title={Boundary refinement network for colorectal polyp segmentation in colonoscopy images},
  author={Yue, Guanghui and Li, Yuanyan and Jiang, Wenchao and Zhou, Wei and Zhou, Tianwei},
  journal={IEEE Signal Processing Letters},
  volume={31},
  pages={954--958},
  year={2024},
  publisher={IEEE}
}

@article{alom2018recurrent,
  title={Recurrent residual convolutional neural network based on u-net (r2u-net) for medical image segmentation},
  author={Alom, Md Zahangir and Hasan, Mahmudul and Yakopcic, Chris and Taha, Tarek M and Asari, Vijayan K},
  journal={arXiv preprint arXiv:1802.06955},
  year={2018}
}

@article{sun2026iseg,
  title={iseg: An iterative refinement-based framework for training-free segmentation},
  author={Sun, Lin and Cao, Jiale and Xie, Jin and Khan, Fahad Shahbaz and Pang, Yanwei},
  journal={IEEE Transactions on Pattern Analysis and Machine Intelligence},
  year={2026},
  publisher={IEEE}
}

@article{kendall2017uncertainties,
  title={What uncertainties do we need in bayesian deep learning for computer vision?},
  author={Kendall, Alex and Gal, Yarin},
  journal={Advances in neural information processing systems},
  volume={30},
  year={2017}
}

@inproceedings{ronneberger2015u,
  title={U-net: Convolutional networks for biomedical image segmentation},
  author={Ronneberger, Olaf and Fischer, Philipp and Brox, Thomas},
  booktitle={International Conference on Medical image computing and computer-assisted intervention},
  pages={234--241},
  year={2015},
  organization={Springer}
}

@article{chen2021transunet,
  title={Transunet: Transformers make strong encoders for medical image segmentation},
  author={Chen, Jieneng and Lu, Yongyi and Yu, Qihang and Luo, Xiangde and Adeli, Ehsan and Wang, Yan and Lu, Le and Yuille, Alan L and Zhou, Yuyin},
  journal={arXiv preprint arXiv:2102.04306},
  year={2021}
}

@inproceedings{lin2017focal,
  title={Focal loss for dense object detection},
  author={Lin, Tsung-Yi and Goyal, Priya and Girshick, Ross and He, Kaiming and Doll{\'a}r, Piotr},
  booktitle={Proceedings of the IEEE international conference on computer vision},
  pages={2980--2988},
  year={2017}
}

\end{document}